\documentclass[10pt,twocolumn,letterpaper]{article}

\usepackage{wacv}
\usepackage{times}
\usepackage{epsfig}
\usepackage{graphicx}
\usepackage{amsmath}
\usepackage{amssymb}
\usepackage{color}
\usepackage{subfigure}
\usepackage{times}
\usepackage{epsfig}
\usepackage{booktabs}
\usepackage{array}
\usepackage{rotating}

\wacvfinalcopy 

\def\wacvPaperID{****} 

\ifwacvfinal\fi
\begin{document}

\title{LaneNet: Real-Time Lane Detection Networks for Autonomous Driving}

\author{Ze Wang\thanks{The majority of the work was done while interning at Horizon Robotics, Inc..}\\
Department of Electrical and Computer Engineering, Duke University\\
{\tt\small ze.w@duke.edu}
\and
Weiqiang Ren \\
Horizon Robotics, Inc.\\
{\tt\small weiqiang.ren@hobot.cc}
\and
Qiang Qiu \\
Department of Electrical and Computer Engineering, Duke University\\
{\tt\small qiang.qiu@duke.edu}
}
\maketitle
\ifwacvfinal\thispagestyle{empty}\fi

\begin{abstract}
	
	Lane detection is to detect lanes on the road and provide the accurate location and shape of each lane. It severs as one of the key techniques to enable modern assisted and autonomous driving systems. 
	However, several unique properties of lanes challenge the detection methods. The lack of distinctive features makes lane detection algorithms tend to be confused by other objects with similar local appearance. Moreover, the inconsistent number of lanes on a road as well as diverse lane line patterns, e.g. solid, broken, single, double, merging, and splitting lines further hamper the performance.
	In this paper, we propose a deep neural network based method, named LaneNet, to break down the lane detection into two stages: lane edge proposal and lane line localization. Stage one uses a lane edge proposal network for pixel-wise lane edge classification, and the lane line localization network in stage two then detects lane lines based on lane edge proposals.
	Please note that the goal of our LaneNet is built to detect lane line only, which introduces more difficulties on suppressing the false detections on the similar lane marks on the road like arrows and characters.
	Despite all the difficulties, our lane detection is shown to be robust to both highway and urban road scenarios method without relying on any assumptions on the lane number or the lane line patterns. The high running speed and low computational cost endow our LaneNet the capability of being deployed on vehicle-based systems. Experiments validate that our LaneNet consistently delivers outstanding performances on real world traffic scenarios. 
	
\end{abstract}

\section{Introduction}


Optical image based lane detection method is a key component of modern driving assistant systems. 
However, the detection of lanes remains challenging for several reasons. First, the appearance of a lane is typically extremely simple, which provides no complicated or distinctive features for lane detection, and dramatically increases the risk of false positive detections. Furthermore, diverse lane patterns, such as solid, broken, splitting, and merging lanes make independent lane modelling difficult. 
Algorithms based on hand-crafted features can only solve the lane detection in limited scenarios. And most existing methods also require strict assumptions on lanes, e.g. lanes are parallel \cite{Aly2008Real,Jiang2010Computer,Nieto2008Robust,Deusch2012A}, lanes are straight or close to straight \cite{Li2016Lane,Niu2016Robust}, which are not always valid especially in urban situations. In recent year, some methods \cite{Zhao2012A,Hur2013Multi} have been proposed to address lane detection under few assumptions to the lanes, yet there is still a large space for further progress on the robustness to diverse real-world scenarios. Methods based on deep neural networks \cite{Huval2015An,Li2016Deep,Gurghian2016DeepLanes}, especially convolutional neural networks stimulate a promising research direction and also inspire the idea of our LaneNet.
Moreover, considering that the lane detection runs on vehicle-based systems, where computation resources are severely limited, the computational cost of a lane detection method should also be considered as a key indicator of the overall performance.

In this paper, to strive for a generalized, low computational cost, and real-time vehicle-based solution, we propose a lane detection method called LaneNet. 
The proposed LaneNet breaks down the lane detection task into two stages, i.e. lane edge proposal and lane line localization, respectively; and each involves an independent deep neural network. In the lane edge proposal stage, the proposal network runs binary classification on every pixel of an input image for generating lane edge proposals, which are served as the input to the lane line localization network in the second stage.

The neural networks in both stages are designed for high accuracy, low computational cost, and high running speed. 
Specifically, a light-weight encoder-decoder architecture is adopted for lane edge proposal, where stacked depthwise separable
convolution and $1 \times 1$ convolution layers are used for fast feature encoding, and non-parametric decoding layers for fast feature
resolution recovery.
The obtained proposal map is then transformed to lane edge coordinates and fed to the second stage, where, a high-speed lane line localization network, consisting of  a point feature encoder and a LSTM decoder, localize the lane lines robustly under various scenarios.


Such two-stage design of LaneNet brings additional desirable properties. Firstly, the lane edge map produced by the proposal network serves as interpretable intermediate features, which to some extent alleviates the impact of the black-box property of neural network based method, and makes the detection failures more trackable. The two-stage process allows the parameters of lane line localization network to be refined in a weakly-supervised manner which alleviate the strong demand for well-annotated training samples.
Additional, an efficient dimensionality reduction is performed when transforming the lane edge proposal map to the lane edge coordinates between the two stages, which further reduces the complexity and the network scale of the lane line localization network, and speed up the detection without any compromise to the accuracy.
Last but not the least, the function of lane edge proposal network can be integrated into semantic segmentation network and further reduces the overall computational cost of the driving assistant systems.

All these advantages mentioned above contribute to a robust and reliable lane detection for autonomous driving systems and driving assistant systems. Extensive experiments as well as comparisons are conducted to validate our LaneNet as a lane detection method with remarkable accuracy and fast speed.

The remaining of this paper is structured as follows: 
In Section~\ref{sec:model} we concisely describe the architecture of our model. We provide the details regarding both the training data and the training process in Section~\ref{sec:train}. Experiments and results are shown in Section~\ref{sec:exp}. Section~\ref{sec:rw} provides a brief overview and summary on other lane detection methods. We conclude the paper with Section~\ref{sec:conclu}.

%


\section{Lane proposal and localization networks}
\label{sec:model}
In this section, we first concisely describe two neural networks designed to together accomplish lane detection in diverse scenarios. The first proposal network detects the edges of lane marks, and produce a pixel-wise lane edge proposal map. The second localization network determines the localization and shape of each lane based on lane edge proposals.
Follows the work of \cite{Aly2008Real}, our LaneNet only takes the Inverse Perspective Mapping (IPM) of an image as input, which helps get rid of the perspective effect in the image, and so lanes that appear to converge at the horizon line are now vertical and parallel. However, our LaneNet does not rely on the assumption that the lanes are parallel (or close to parallel) so that the robustness is guaranteed.
The two proposed networks play different roles in lane detection, yet they share several common design principles. Firstly, different from those algorithms running on servers or even computation centers, lane detection runs on vehicles so that the computational cost should be reduced as possible while the inference should be fast enough for real-time decision making. Additionally, the prediction accuracy and the robustness to diverse scenarios are crucial. The details of our LaneNet and how to meet the aforementioned principles are discussed in this section.

\subsection{Lane edge proposal network}


\begin{figure*}[t]
	\begin{center}
		\subfigure[Lane edge proposal network. The layers of encoder and decoder are marked in blue and red, respectively.]{
			\begin{minipage}{0.5\linewidth}
				\includegraphics[width=1.0\linewidth]{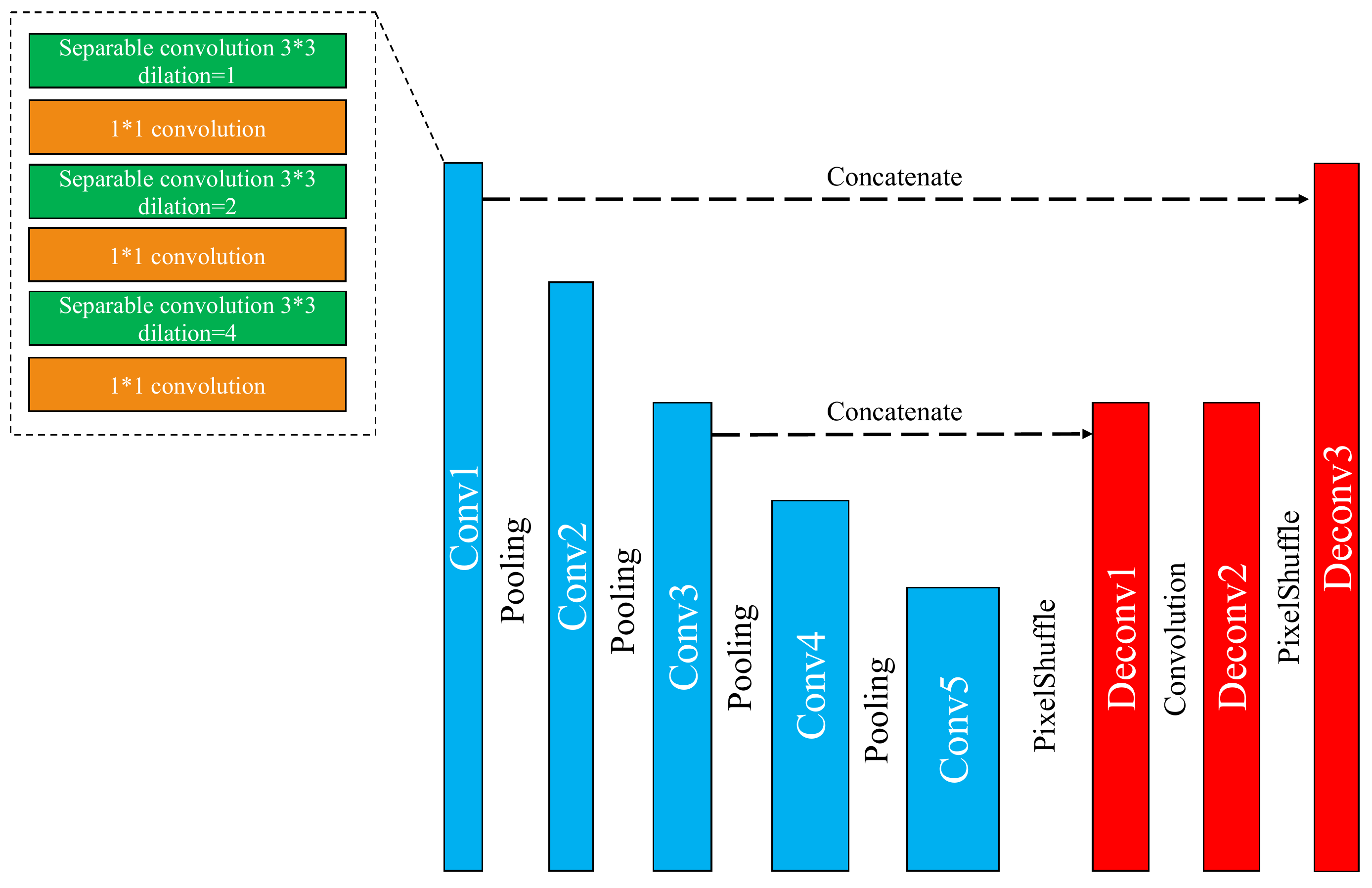}
		\end{minipage}}
		\subfigure[Lane line localization network.]{
			\begin{minipage}{0.7\linewidth}
				\includegraphics[width=1.0\linewidth]{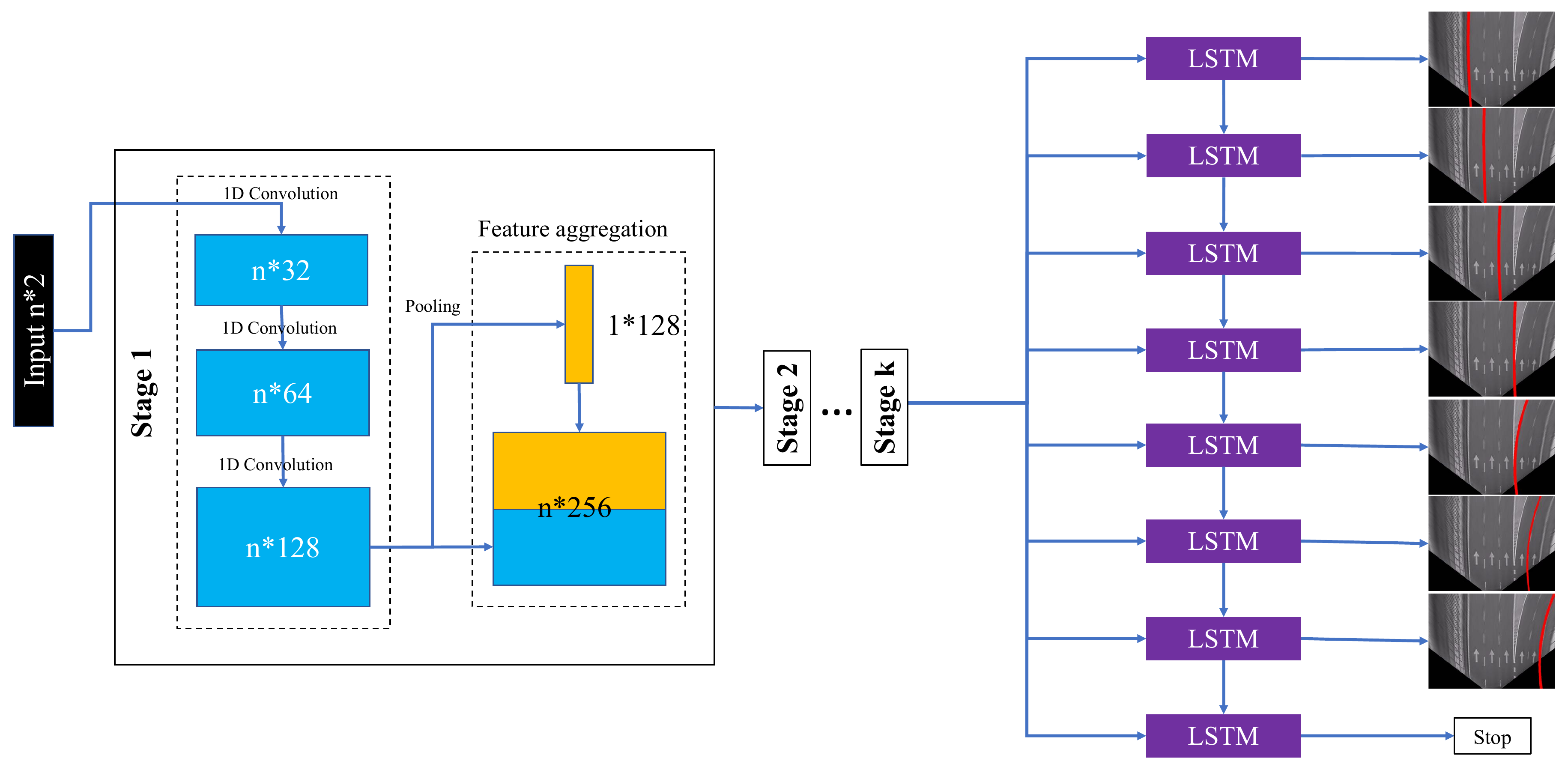}
		\end{minipage}}
	\end{center}
	\caption{The architectures of the lane edge proposal network in the first stage and the lane line localization network in the second stage.}
	\label{fig:arc}
\end{figure*}
\begin{figure}[t]
	\begin{center}
		\includegraphics[width=1.0\linewidth]{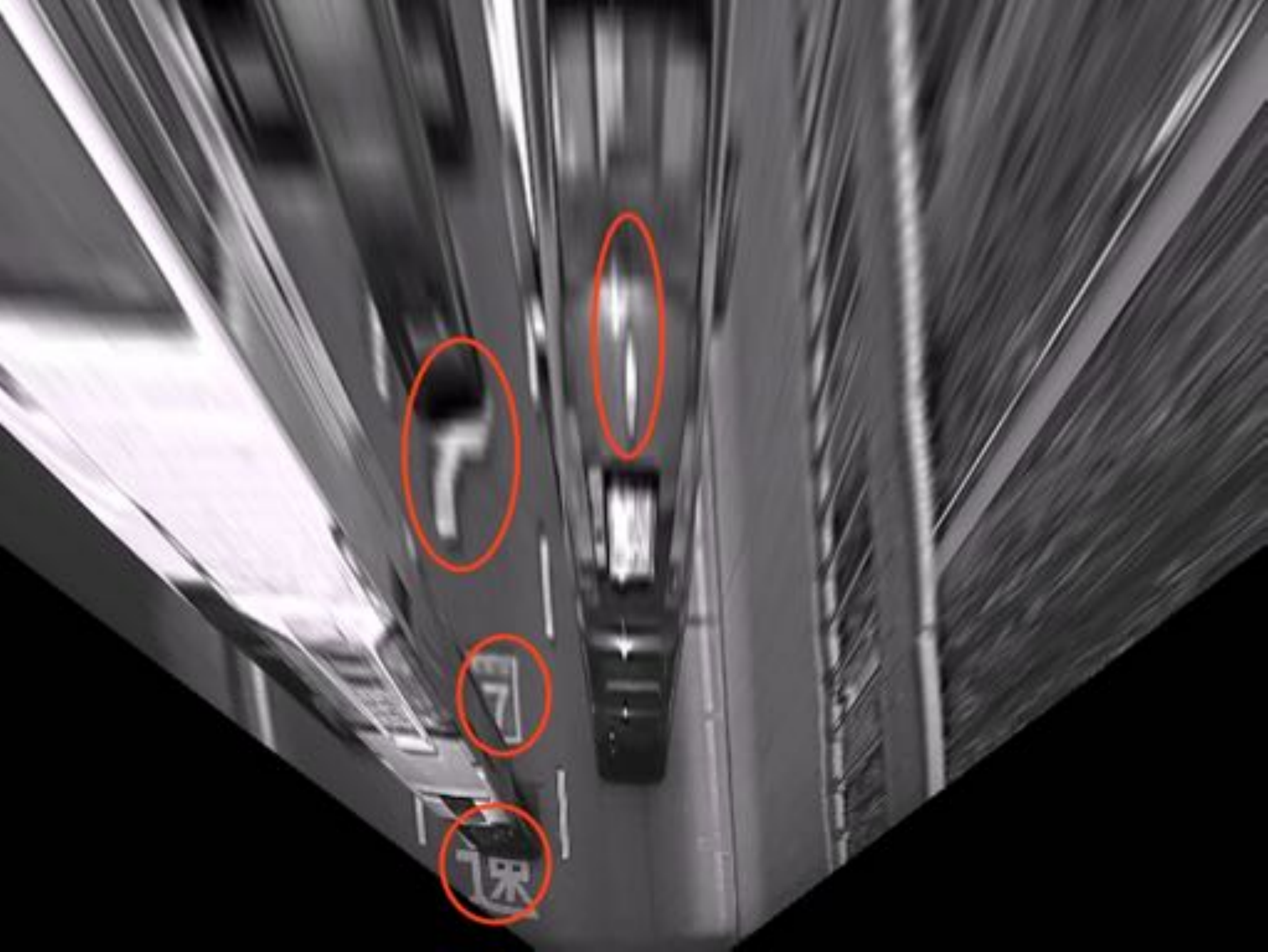}
	\end{center}
	\caption{A typical IPM image of the front view of a vehicle. The red circles marks the areas that are easy to be detected as false positives.}
	\label{fig:ipm}
\end{figure}

To filter irrelevant information contained in optical images, we first process an image by a lane edge proposal network to generate a binary lane edge proposal map indicating the position of the pixels likely lie on the edge of lane segments. 
The reason we adopt a deep neural network instead of using hand-craft features to proposal maps is that it is essentially difficult to find one or a set of robust hand-craft features for filtering lane edges in an input image. As an illustration, Fig.~\ref{fig:ipm} shows a typical IPM image of the front view of a vehicle. As marked by red circles, road marks, characters on the roads, guardrails along the roads, and even the light reflected by other vehicles share similar local appearances with lanes, thus are easy to be detected as false positives. To derive accurate lane edge proposal on each pixel, contextual information as well as global appearance should all be taken into consideration to suppress false positive detections caused by similar local appearances. Deep convolutional neural network thus become a desirable tool thanks to its powerful feature extraction capacity.


Encoder-decoder architectures are widely used in dense prediction tasks like semantic segmentation \cite{Long2014Fully,Ronneberger2015U,Badrinarayanan2017SegNet}, which typically utilize convolutional layers and deconvolutional layers (transpose convolution layers) for feature encoding and decoding.
For a higher efficiency, our lane edge proposal network adopts a light-weight encoder-decoder architecture. The encoder takes an IPM image of the front view of a vehicle as the input, and hierarchically extracts the features. The decoder progressively recovers the resolution of the feature map and produce a pixel-wise lane edge proposal map.

\paragraph{Encoder}The encoder of the proposal network 
replaces the standard convolution operations by the combination of depthwise separable convolutions and pointwise convolutions ($1 \times 1$ convolutions) to significantly lower the computational cost as in \cite{Howard2017MobileNets}. Specifically, depthwise separable convolutional layers with a kernel size of 3 are stacked for progressive feature extraction. Each depthwise separable convolution layer is followed by a $1 \times 1$ convolution layer for channel-wise information aggregation. As mentioned above, there exists a lot of objects that share similar local appearance with lanes in the input images, to prevent false positive detections, the context information should be properly extracted and preserved in the encoding stage. In order to enlarge the reception field of the encoder while guarantee the low computational cost, we adopt dilated convolutional kernels in the encoder layers. 
Specifically, as plotted in Fig.~\ref{fig:arc} (a), three depthwise convolution layers that followed by a $1 \times 1$ convolution layer are included to form a convolution block that is used for feature extraction on one particular feature resolution. Among the three depthwise separable convolutional layers, the first one has a dilation rate of 1, which corresponds to a standard separable convolution, while the following two layers use dilation rate of 2 and 4, respectively, to enlarge the reception field. Introducing dilated kernels successfully improves the valid reception field without any additional parameters or computational cost, which properly balances the efficiency and effectiveness.

\paragraph{Decoder}In order to recover the feature resolution and produce the pixel-wise lane edge map of an input image, we design a decoder architecture that follows the encoder. 
Deconvolution layers (transpose convolution layers) are widely used to enlarge the intermediate features in modern deep neural networks. However, the computational cost and the training difficulty make deconvolution layers an inappropriate choice in our model. Inspired by the success of adopting sub-pixel convolution layers in image super-resolution \cite{Shi2016Real}, we use sub-pixel convolution layers here in our model to progressively recover the feature resolution. It brings the beneficial properties including completely parameter-free and no computational cost, which are much desirable in our case of lane detection. Skip connections are used to provide high resolution feature for accurate lane edge localization. 
The final high resolution feature map is then used to generate the pixel-wise lane edge probability map and the value of each pixel indicates the confidence that the network decides this pixel lies on the edge of a lane segment in the input image.

\subsection{Lane line localization network}

The lane edge proposal network projects an image to the corresponding binary lane edge proposal map where structure and appearance are far simpler, while detecting the lane lines from a lane edge proposal map remains sophisticated. 
The diverse patterns of lanes, e.g., splitting lanes, merging lanes, intersection, and uncertain total number of lanes 
challenge the detection.
We complete our lane line detection method with a light-weight, ultra fast, yet powerful neural network referred to lane line localization network. Our lane line localization network also adopts an encoding-decoder structure. It takes the coordinates of the lane edges as input, and apply a series of 1D convolution and pooling operations to encode an input to a low dimensional holistic representation. A long short-term memory network (LSTM) based decoder is used to progressively decode the representation to the parameters of each lane in the image.

Specifically, given a binary lane edge map, we use coordinates of lane edge points instead of directly inputting the lane edge map to the lane line localization network. Such a design brings us two major benefits: First is that the size of the input is reduced from $w \times h$ to $n \times 2$, where n is the number of the lane edge points in the binary lane edge map. This reduction allows the compact architecture of our model and the ultra fast prediction. Secondly, such coordinate from of inputs enable a weakly supervised training method of the lane line localization network, which will be discussed in section~\ref{sec:train_c}.

Apart from the benefits, using coordinates, instead of lane edge map as model inputs comes with an additional requirement to the model, which is the invariance to the input order. Since the lane edge points are sampled from 2D images, it's hard to find a proper and reasonable rule to sort the points. And considering that in most cases, the positions of the lane edge points are sufficient to determine the number and the shape of lanes on the road, the prediction of our model should be invariant to the input point order so that the prediction remains consistent regardless any shuffle to the order of the input points. 
Inspired by the recent progress on novel deep neural network architecture PointNet which is specifically designed for analysing point cloud data, an input order insensitive encoder is designed in our lane line localization network to encode the input points to a low dimensional holistic representation of the lane edges.
Specifically, to achieve invariance to the input order, shared fully-connected layers are utilized to perform non-linear projection on the feature of each point simultaneously. This operation can be simply implemented by a series of 1D convolutional layers whose kernel size is identical to the size of the feature of each point. Max pooling layer is adopted to aggregate the feature of all the points to achieve a global information fusion which is also insensitive to the input order. Different from the model proposed in PointNet, we do not use any transformation to the input or the features, and the max pooling layer is used for multiple times to encourage the information communication among all the points under multiple semantic levels. Specifically, we refer a series of non-linear feature extraction to each point and a pooling based feature aggregation to a stage in encoder network as shown in Fig.~\ref{fig:arc} (b), and multiple stages are used in our network for effective information communication among all the points and the final representation is drawn by applying average pooling to the final feature.

After the input coordinates of the points are encoded to a low-dimensional representation, we adopt LSTM, which serves as a solution to the uncertain number of lanes, to decode the parameters of each lane. Specifically, we fit each lane in the image by a quadratic function, and use LSTM to progressively predict the parameters depicting each lane along with a confidence score from the left side of the image to the right side of the image one by one. The prediction is terminated when the confidence score is lower than a predefined threshold which indicates the network believes there is no more lane in the image.

\section{Training}
\label{sec:train}

\subsection{Lane edge proposal network}

The training of the lane edge proposal network is straightforward. In our case, the model takes an image of the front view of a vehicle as input, and outputs a lane edge probability map of the same size as the input image. All the parameters are trained in a completely supervised manner. For each training image, we provide an annotation map where '1' indicates this pixel is a positive point laying on the edge of one lane segment, and '0' elsewhere. The entire network is end-to-end optimized by stochastic gradient descent. Note that in every annotation map, the number of positive points are always far smaller than number of negative points. So in the training process, we dynamical weight the loss of positive points and negative points according to their relative amounts, and the loss function is written as:
\begin{equation}\label{}
l_{cls}= \sum_{i}^{B} \sum_{n}^{k} y_{n}^{i} \times \log P_{n}^{i} + \beta (1 - y_{n}^{i}) \times \log(1 - P_{n}^{i})
\end{equation}
where $B$ is the batch size of one iteration in training process and $P^{i}$ denotes the prediction of the $i$-th sample in this mini-batch. $P_{n}^{i}$ indicating the $n$-th pixel in the image, and $k$ is total number of pixel in the image which equals to $w \times h$. $\beta$ is the ratio of the number of positive points to the number of negative points, which acts as a dynamic weight for balancing the losses from negative points and positive point. It could be computed by 
\begin{equation}\label{}
\beta = \frac{\sum y^{i}}{w \times h - \sum y^{i}}
\end{equation}
for simplicity.

After training, the network directly produces a lane edge proposal map of the input image where the value of each pixel indicates the confidence of this pixel lies on the edge of one lane segment. 

\subsection{Lane line localization network}
\label{sec:train_c}

\begin{figure}[t]
	\begin{center}
		\includegraphics[width=1.0\linewidth]{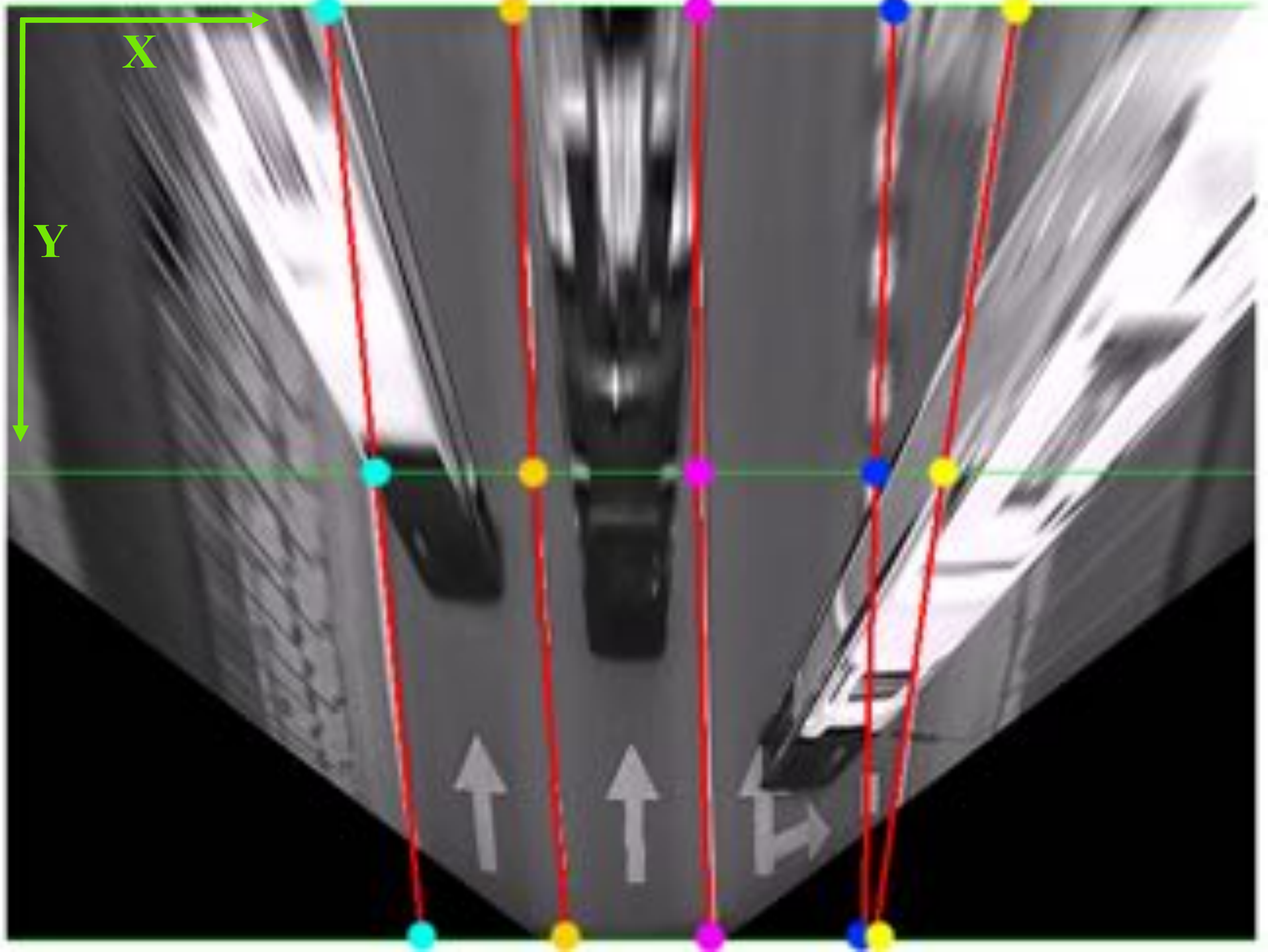}
	\end{center}
	\caption{We present an illustration on how we derive the training objectives of the network. The green line indicates the middle line of the image. We use three points to locate a lane which is fitted using a quadratic function.}
	\label{fig:anno}
\end{figure}

In our data, each lane line is represented as a quadratic function, i.e., the shape of each lane is annotated by three continuous numbers. 
In practice, however, training the network to predict the parameters of each quadratic function works poorly as each parameter is usually of significant different orders of magnitude, e.g. the quadratic term of a lane is nearly zero yet the constant term is thousands times larger, which makes the construction of a loss function difficult. Simply normalize the parameters results in disappointing performance as well. 
Therefore, we solve the problem by training the network
to predict the point locations where a lane intersects with the top, middle and the bottom line of an image, respectively.
Fig.~\ref{fig:anno} shows the way we construct the training objectives for each lane. In our case, the input image has a size of $h \times w$, so the values we train the network to predict are the three $X$ coordination values of the points which lies on the intersections of the lane with three horizontal lines $Y=0$, $Y=h/2$, and $Y=h$. We refer these three values to the keys values of a lane. By transferring the training objective from the parameters of a quadratic function to the key values, the predicted values are of relatively similar orders of magnitude so that the training becomes stable, and converging at bad local minimum is prevented.
A simple matrix multiplication can directly transform the parameters of a quadratic function to the corresponding key values: 

\begin{equation}\label{}
\left [ K_{1}, K_{2}, K_{3} \right ]=\left [ P_{2}, P_{1}, P_{0} \right ] \begin{bmatrix} 0& \left ( h/2 \right ) ^{2} &h^{2}  \\0 & h/2 & h \\0&0&0\end{bmatrix} ,
\end{equation}
where we use $\left [K_{1}, K_{2}, K_{3} \right ]$ to denote the key values of a lane, and $\left [P_{2}, P_{1}, P_{0} \right ]$ to indicate the list of quadratic term, linear term and constant term. 

While training the lane line localization network in a completely supervised way, we can use the average of the $l_2$ distance between predicted key values and the real key values of each lane as the loss function and, minimize the distance using stochastic gradient descent.

Furthermore, using lane edge coordinates as network input brings us an additional important benefit. Considering the large cost of annotating each lane in an image using a quadratic function, and any inaccurate annotations affecting the training result, we design a weakly supervised training method for the lane line localization network, which only requires the number of lanes in a image as ground truth, we call this form of loss the min-distance loss. When using min-distance loss, the loss function is computed by the sum of the distance of each input point to its nearest estimated line.

Specifically, when training the network using min-distance loss, the output of the networks is still the key values of the lanes. After we get the predicted key values, we transform them back to corresponding quadratic functions representing the estimated lanes:
\begin{equation}\label{}
[P_{2}^{pr}, P_{1}^{pr}, P_{0}^{pr}]=[K_{1}^{pr}, K_{2}^{pr}, K_{3}^{pr}]\begin{bmatrix} 0&\left ( h/2 \right ) ^{2}&h^{2}\\0&h/2&h\\0&0&0\end{bmatrix} ^{-1},
\end{equation}
which gives us a set of estimated quadratic functions of all lanes. As mentioned above, the input of the network are the coordinates of all the lane edge points. Now we take all the $X$ coordination values of the points to each estimated quadratic functions, and this will give us the projected $Y$ coordination values of the points to the lanes. The absolute values of the distance between the real $Y$ coordination value and the projected $Y$ coordination values are the horizontal distances of the points to the estimated lanes. Ideally, we take the sum of the horizontal distances of all the points to its nearest estimated lane line as the loss, and minimizing this loss pushes the predicted lane lines to the exact centers of ground truth lane lines. 

This weakly supervised training manner only needs the number of lanes as a supervision so that it significantly reduces the cost of annotation. In practice, only using min-distance loss to train the model from scratch leads to trapping in a poor local minimum thus we have to start with supervised training. 
We come up with two ways to adopt this min-distance loss to the training of our model. First one is that while training the network, we train the network to minimize a combination of the $l_2$ loss and the min-distance loss
\begin{equation}\label{}
\begin{aligned}
L &= L_{l_2} + \alpha L_{min-distance } \\
\end{aligned}
\end{equation}
where $\alpha$ is a hyperparameter for balancing two loss functions. From our observation, minimizing the two loss functions simultaneously successfully overcomes the annotation inaccuracy so that the estimated lines fit exactly to the lanes in the image regardless some minor misalignments between the annotations and the real lanes, and results in an improvement on generalization.

After the network is trained by well-annotated samples, we can further refine the network on some weakly annotated samples which only has the annotation of number of lanes. Since the network has already been optimized, only using min-distance loss for refinement does not collapse the learned model, and can further refines the network parameters.

\begin{figure*}
	\centering
	\subfigure[Original images]{
		\begin{minipage}[b]{0.32\linewidth}
			\includegraphics[width=1\linewidth]{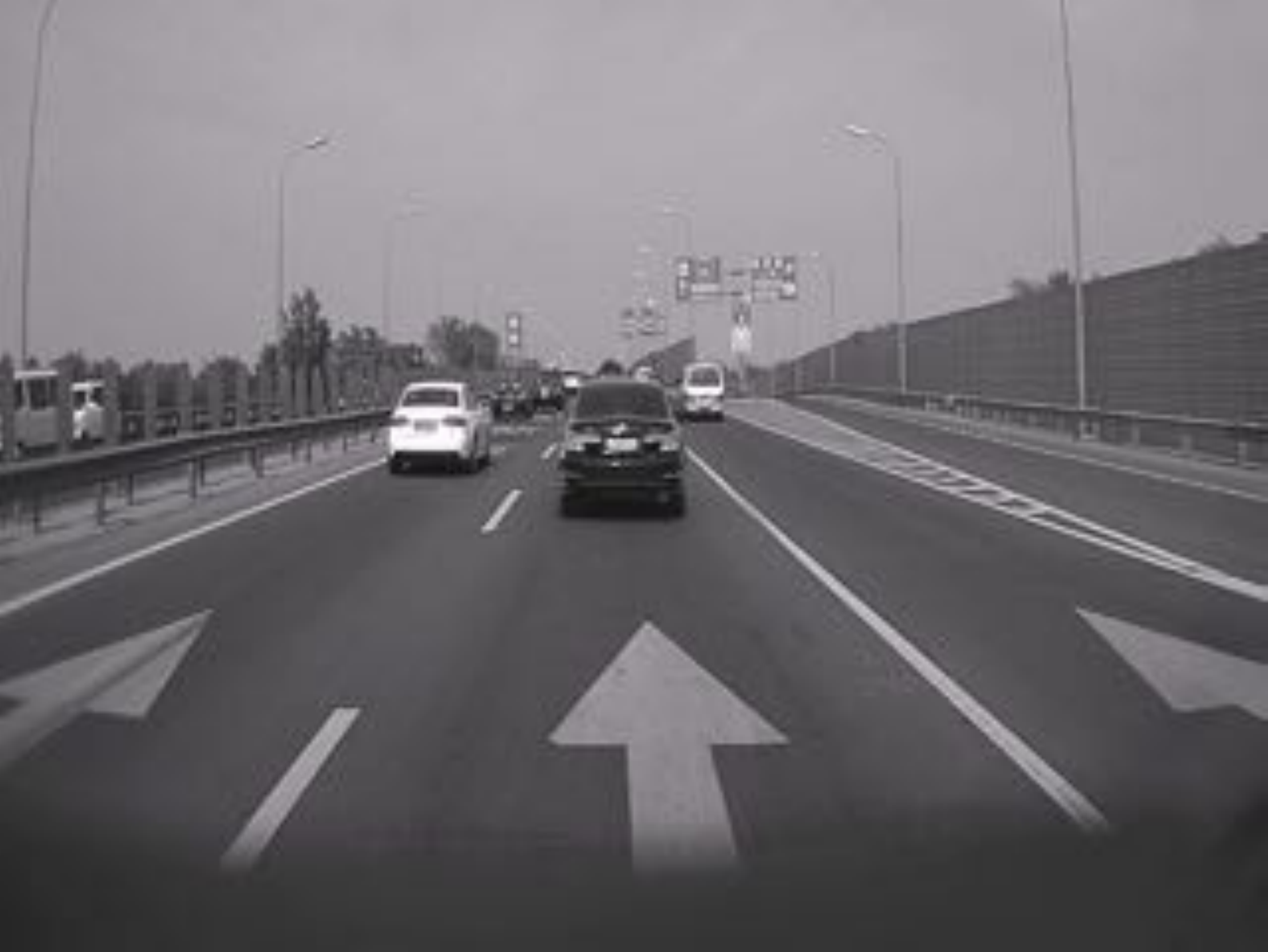}\vspace{5pt}
			\includegraphics[width=1\linewidth]{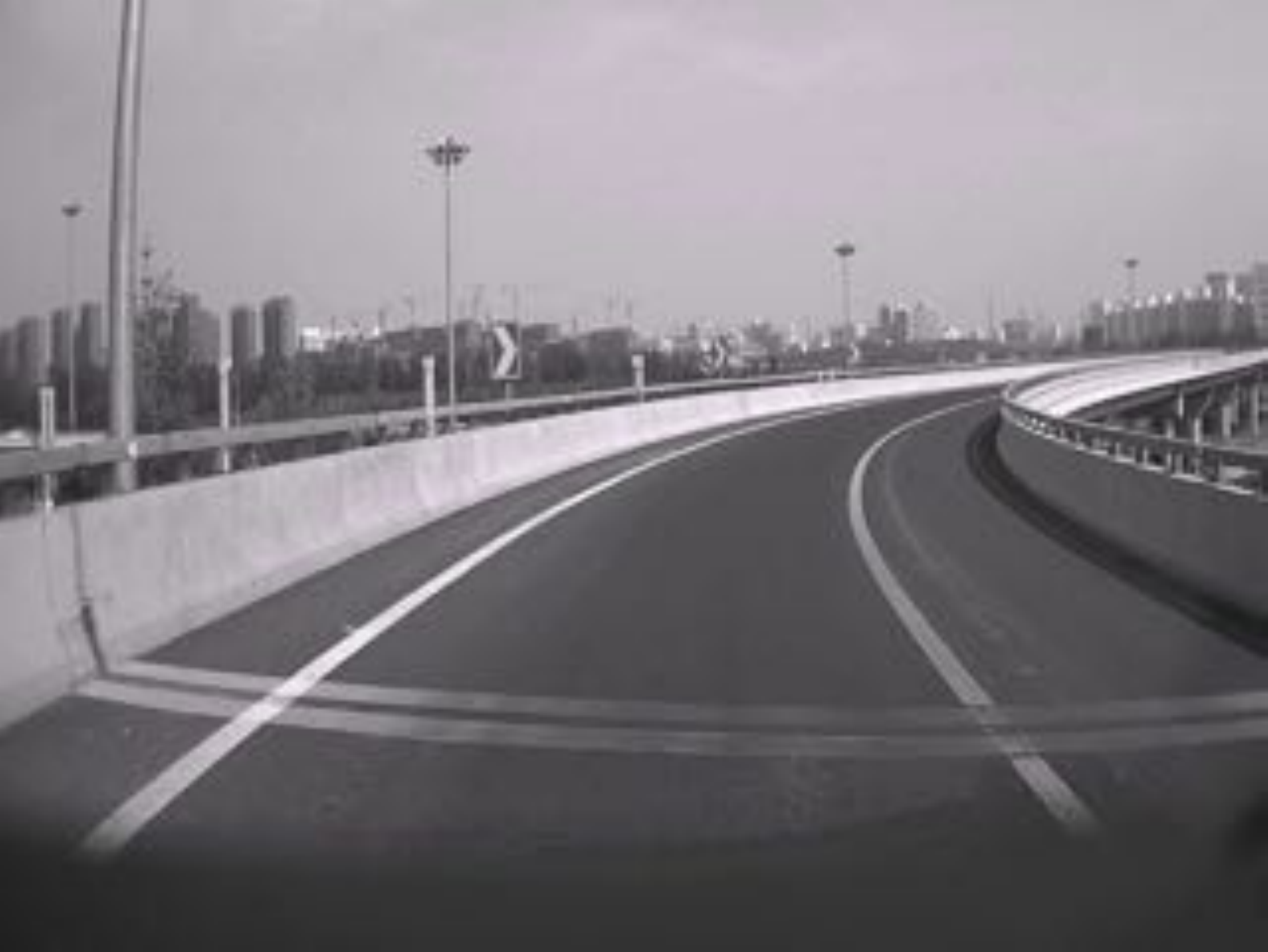}\vspace{5pt}
			\includegraphics[width=1\linewidth]{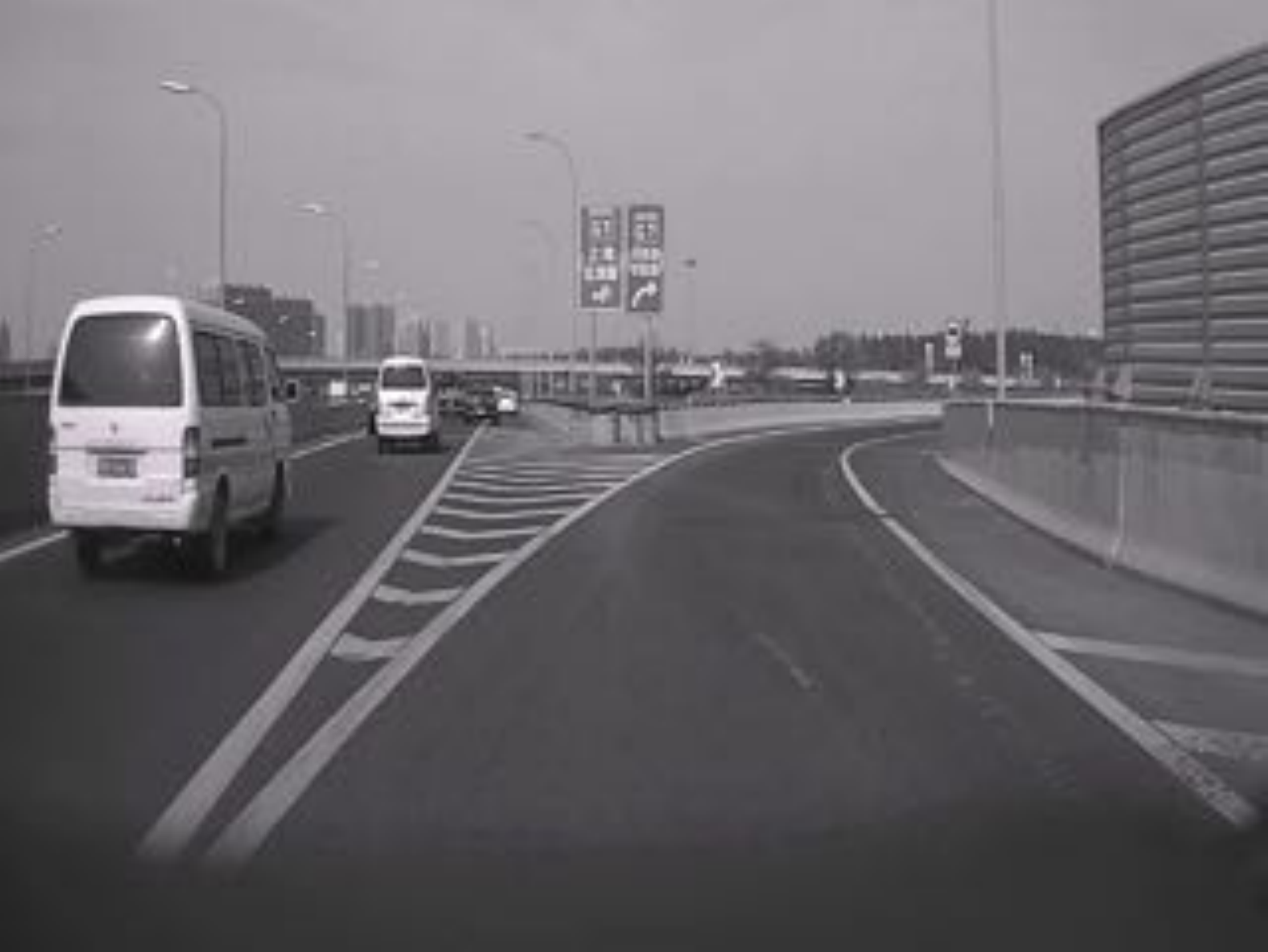}\vspace{5pt}
			\includegraphics[width=1\linewidth]{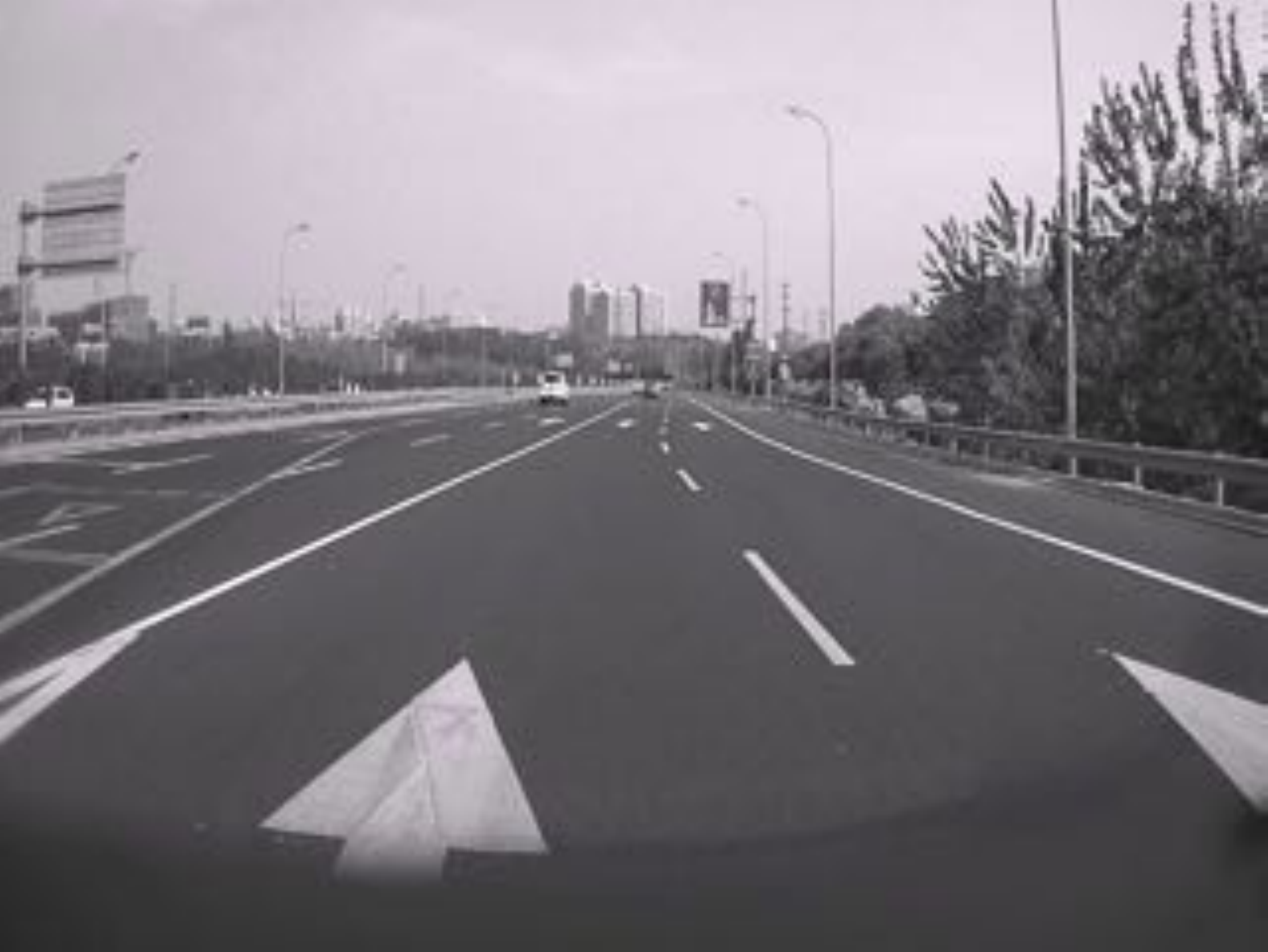}\vspace{5pt}
			\includegraphics[width=1\linewidth]{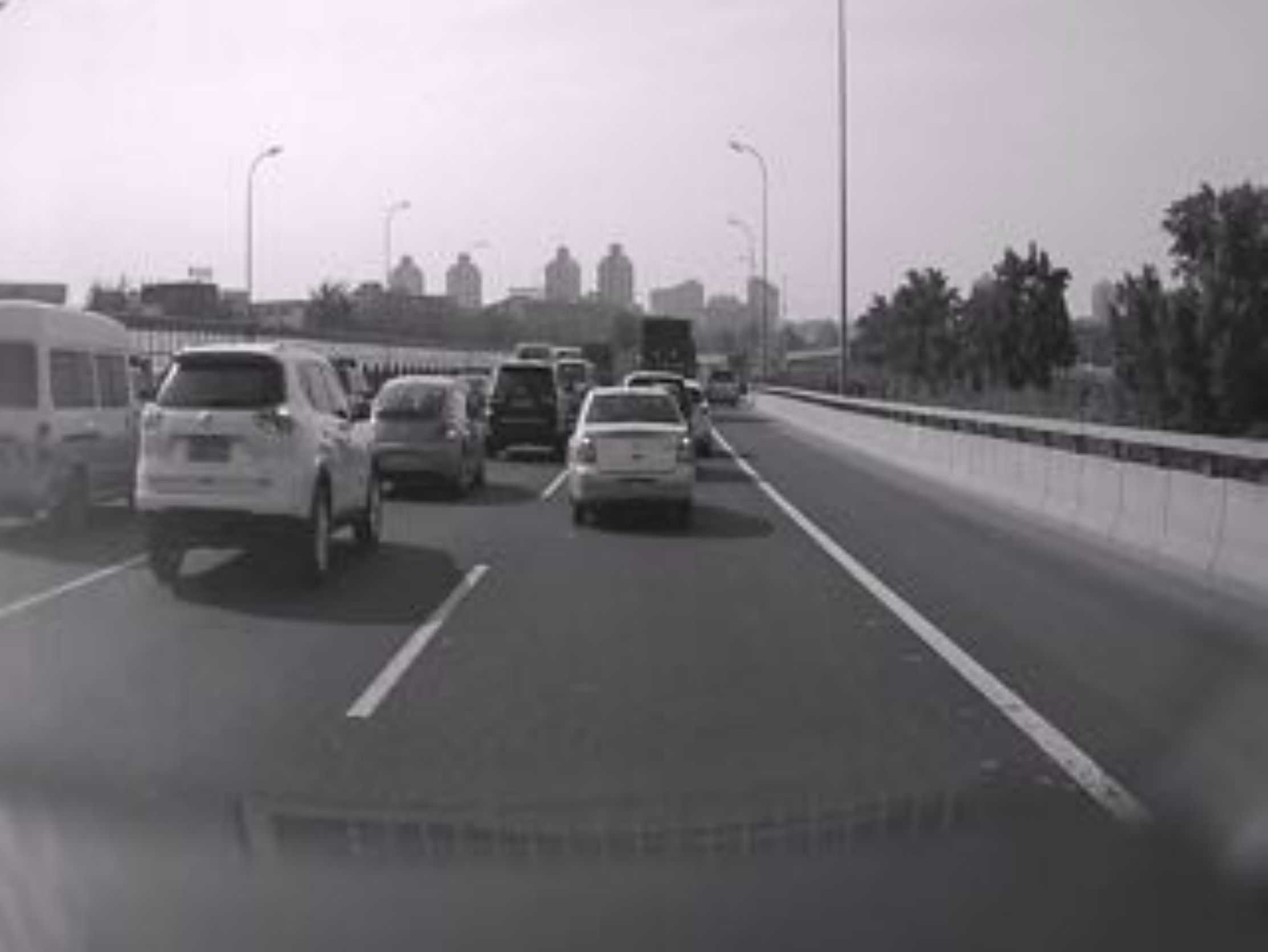}
	\end{minipage}}
	\subfigure[Lane edge proposals]{
		\begin{minipage}[b]{0.32\linewidth}
			\includegraphics[width=1\linewidth]{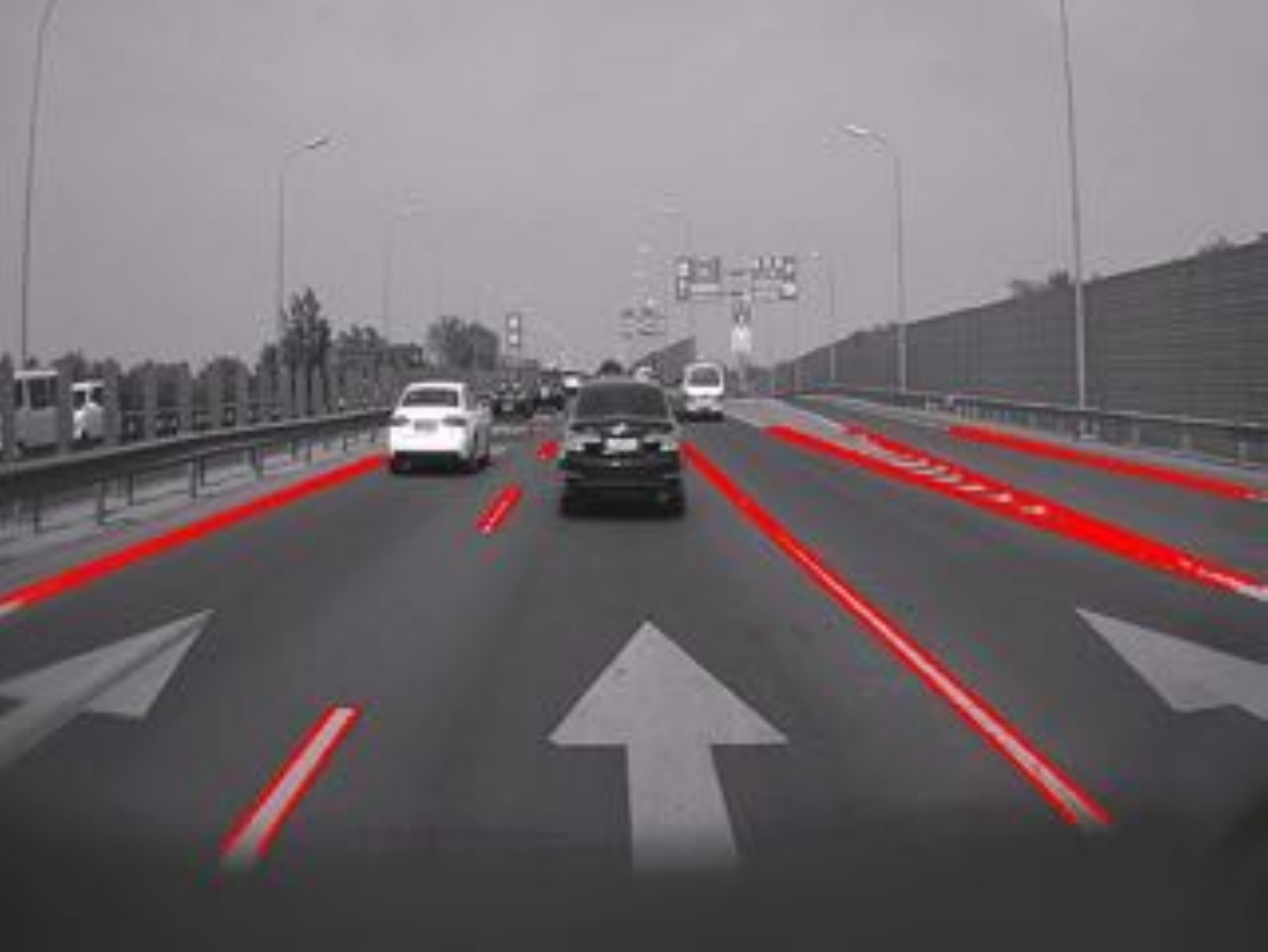}\vspace{5pt}
			\includegraphics[width=1\linewidth]{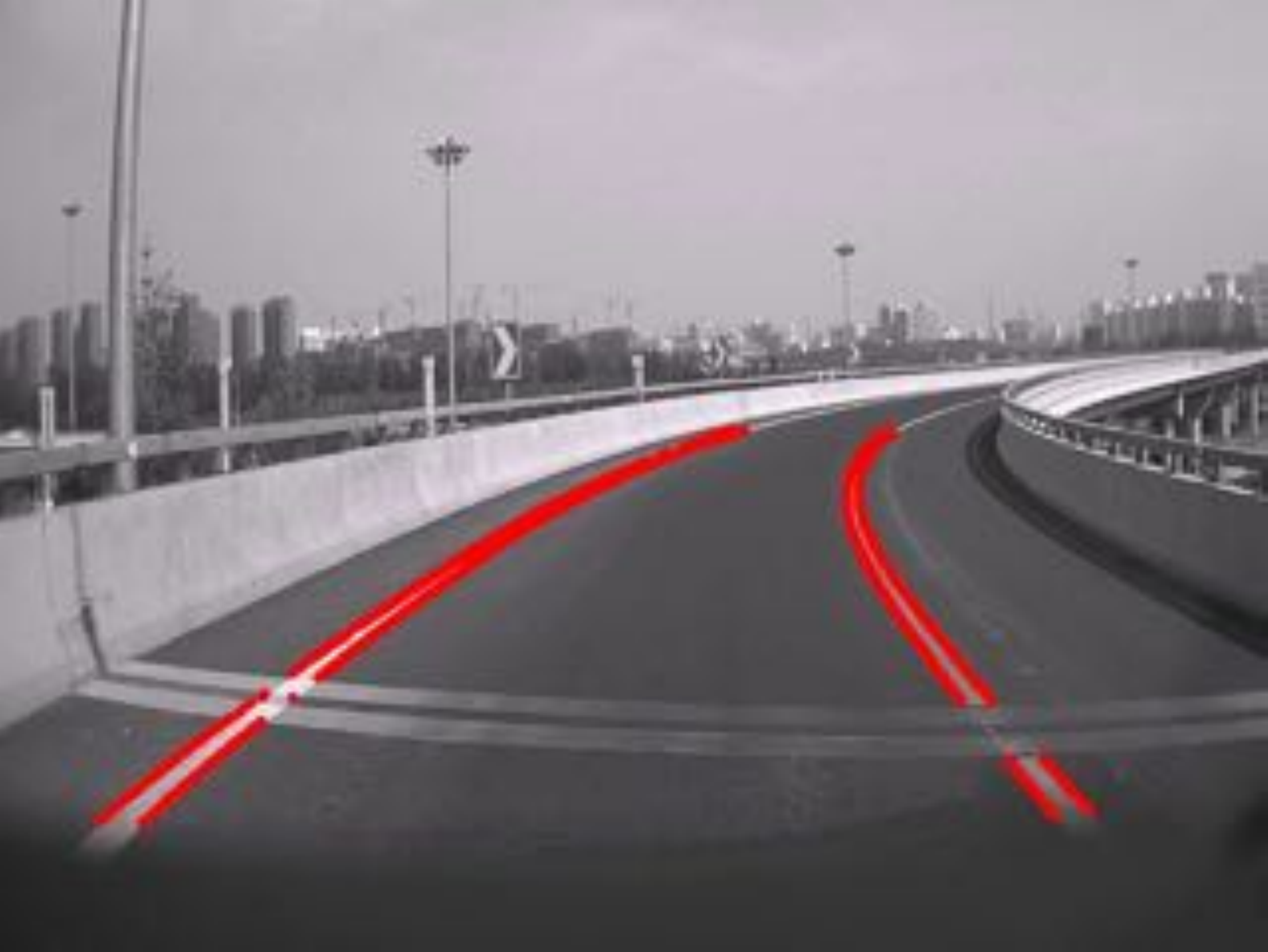}\vspace{5pt}
			\includegraphics[width=1\linewidth]{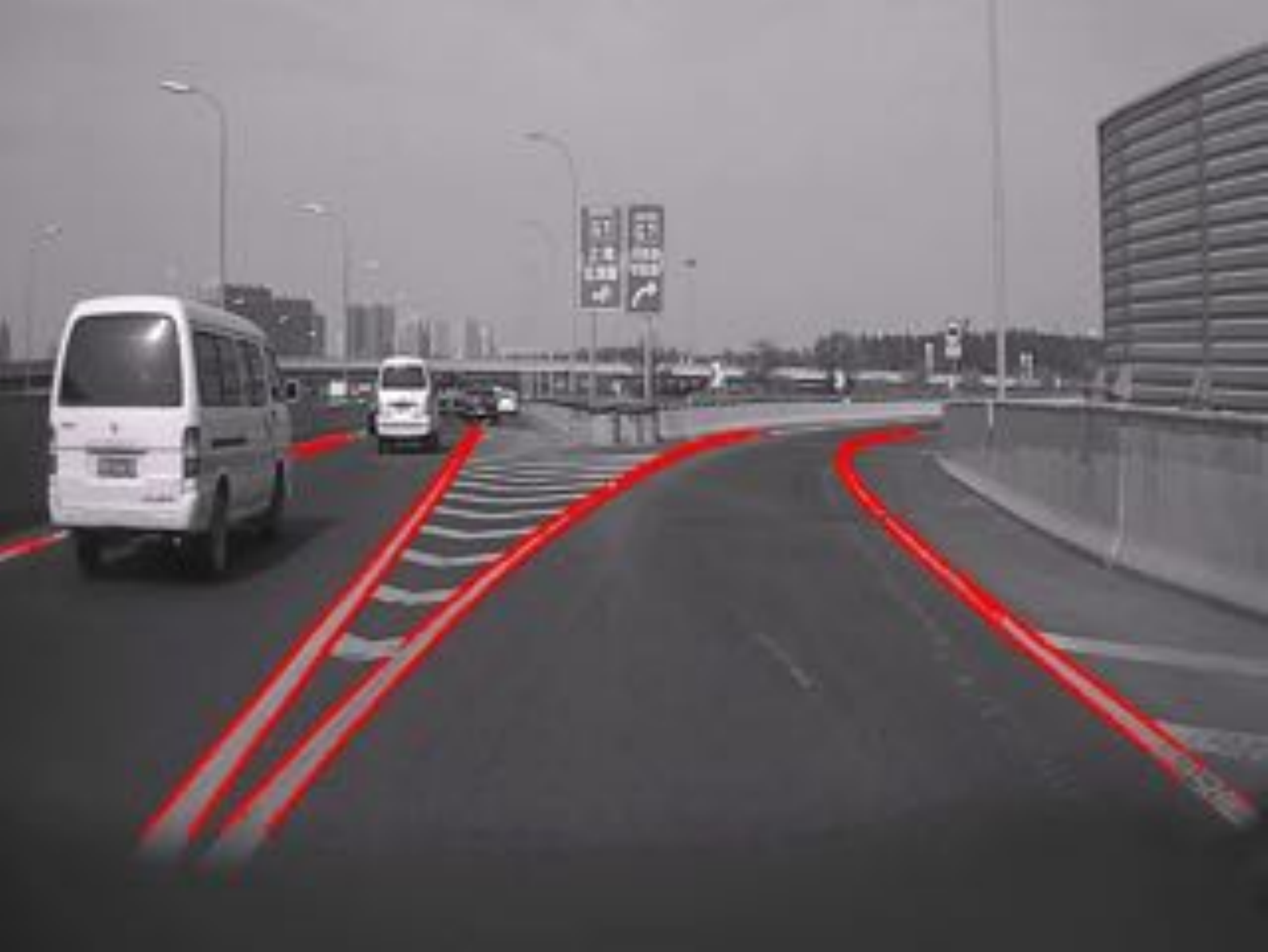}\vspace{5pt}
			\includegraphics[width=1\linewidth]{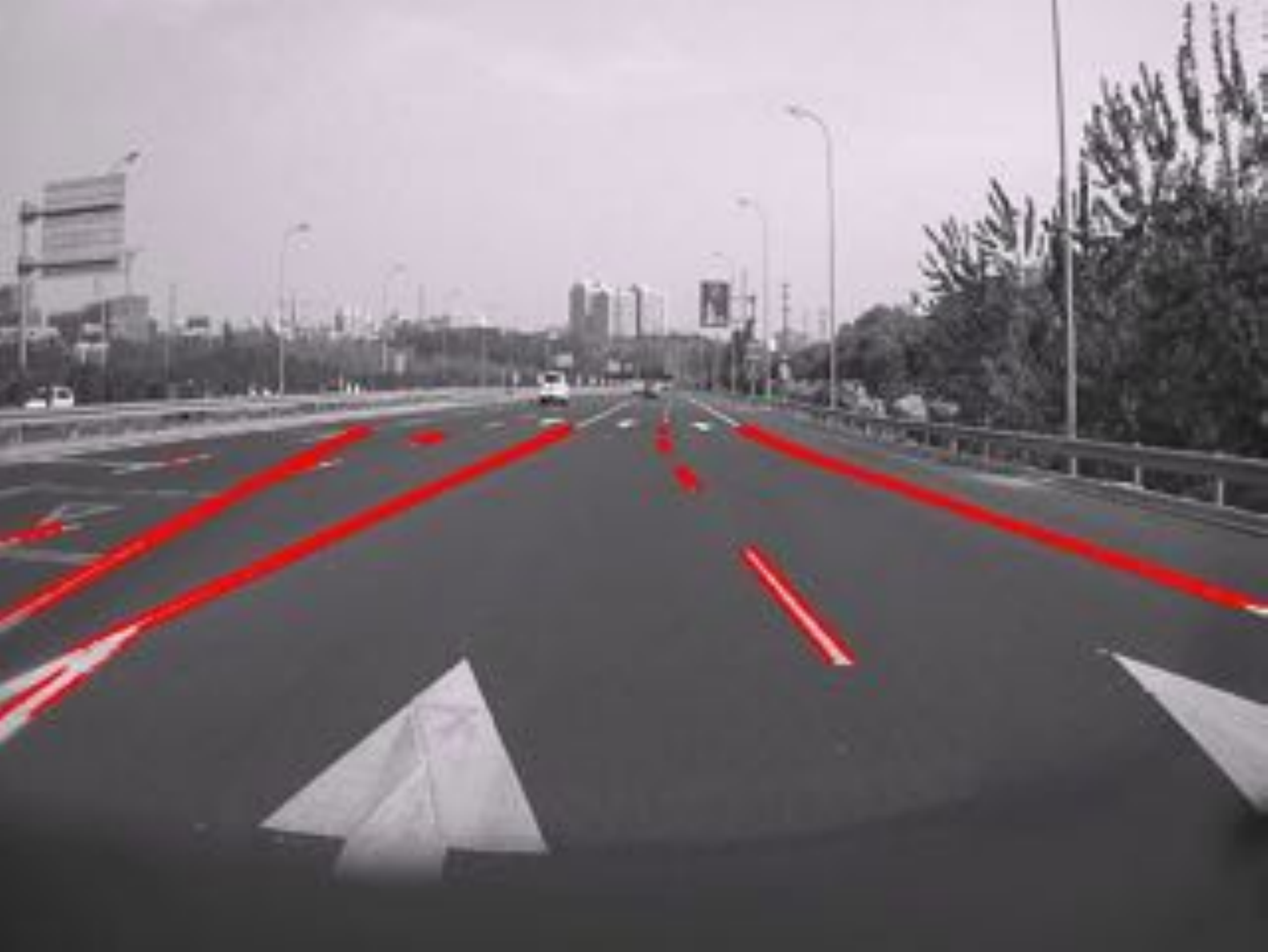}\vspace{5pt}
			\includegraphics[width=1\linewidth]{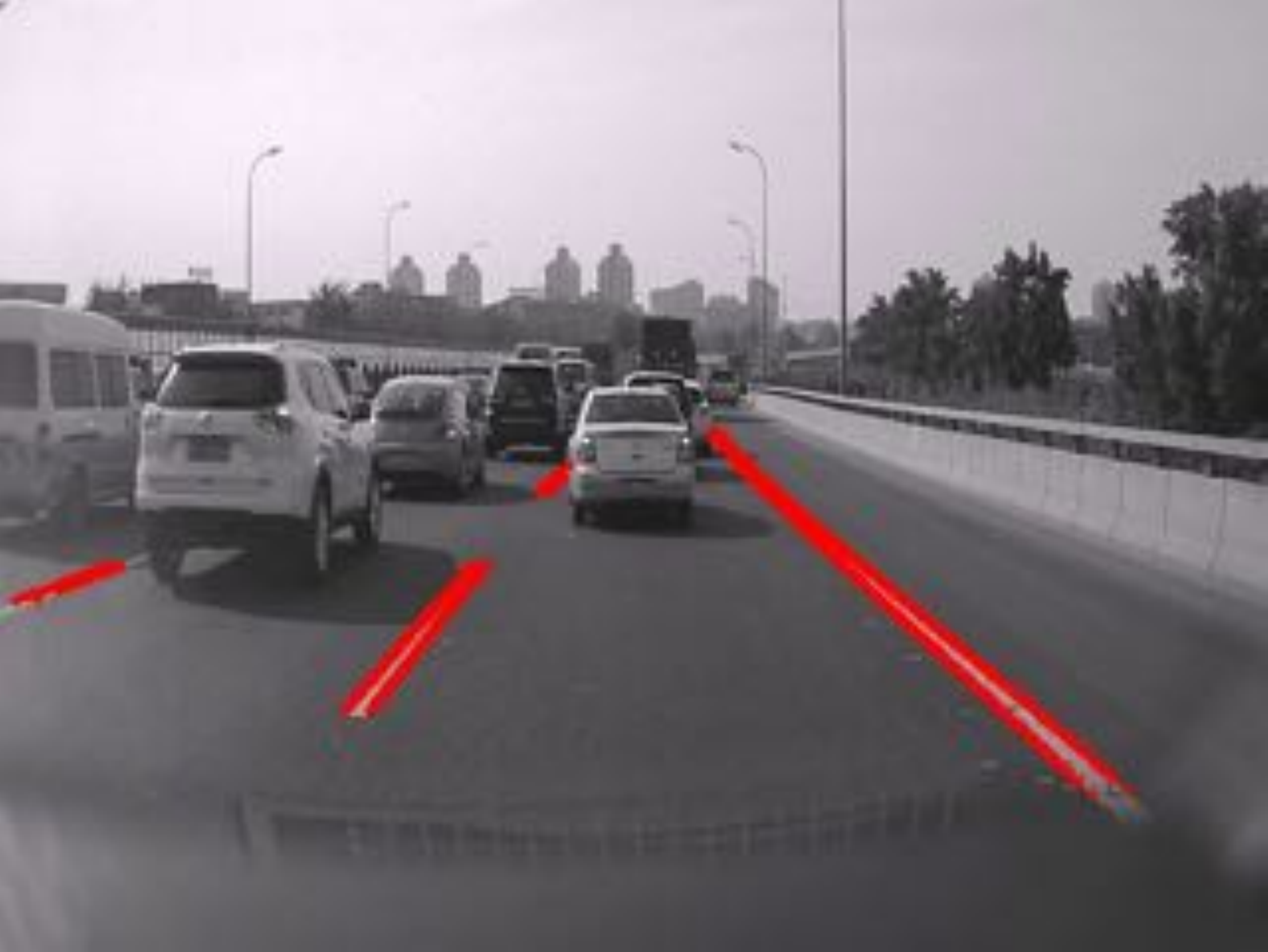}
	\end{minipage}}
	\subfigure[Final results]{
		\begin{minipage}[b]{0.32\linewidth}
			\includegraphics[width=1\linewidth]{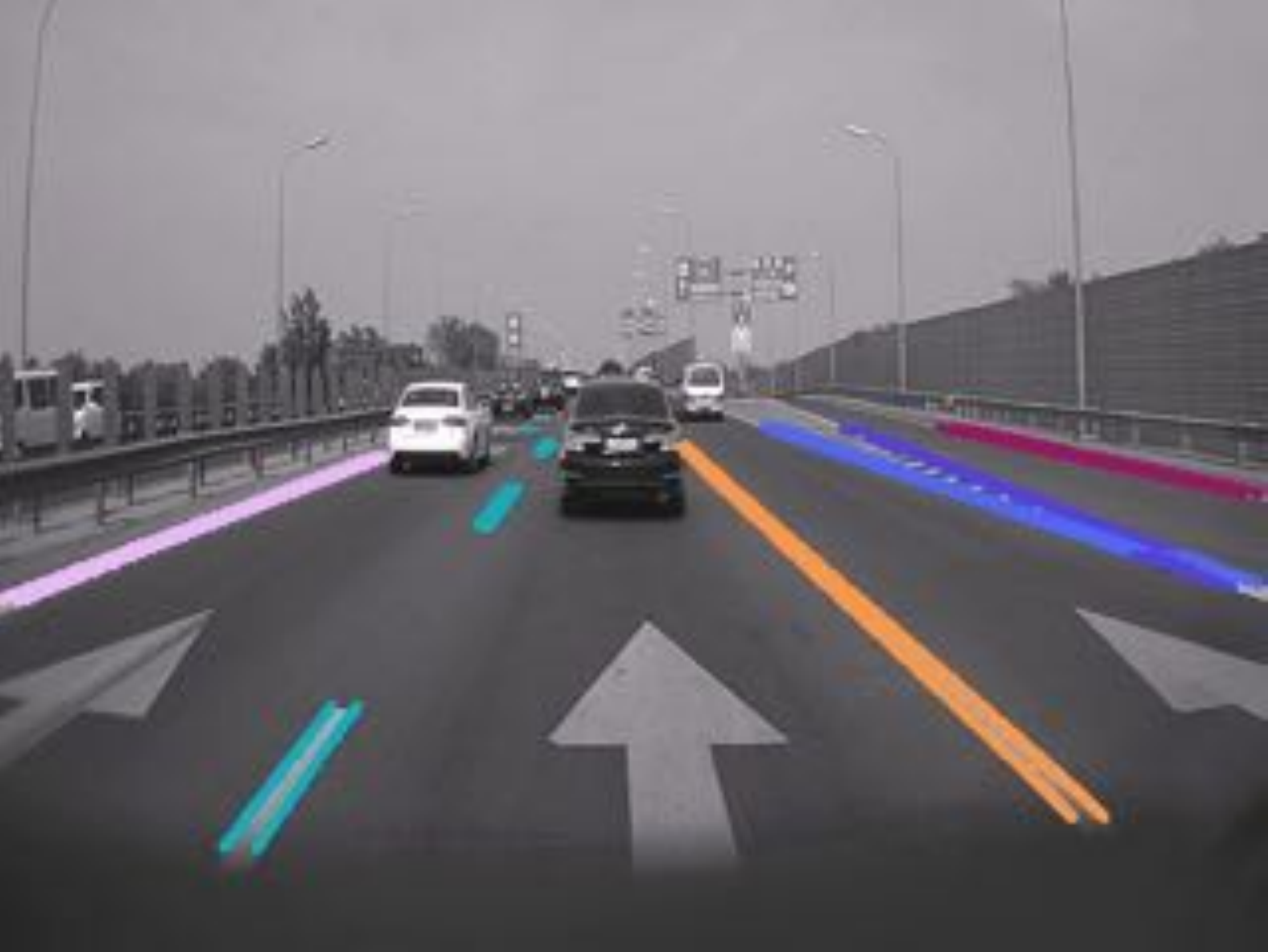}\vspace{5pt}
			\includegraphics[width=1\linewidth]{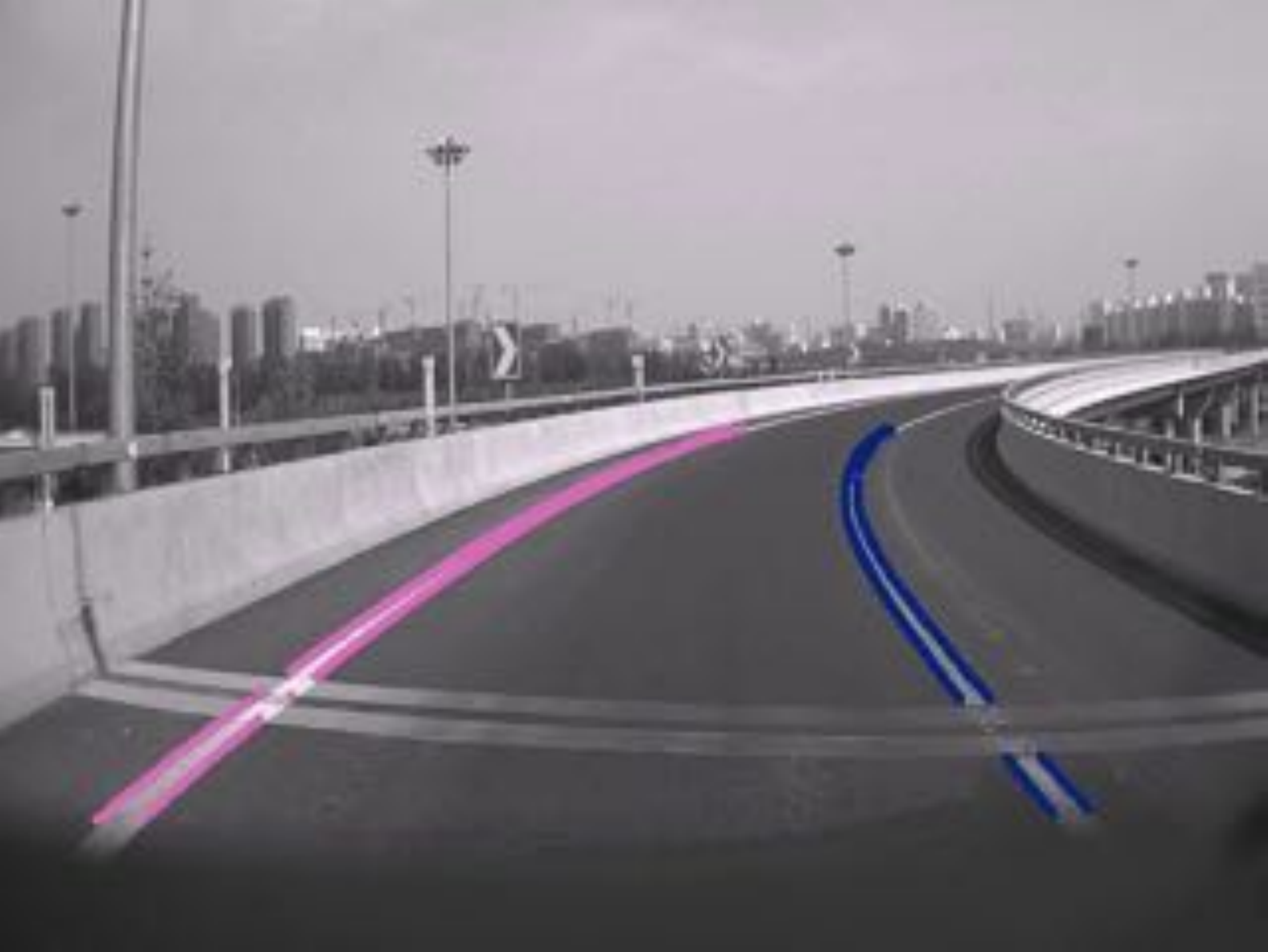}\vspace{5pt}
			\includegraphics[width=1\linewidth]{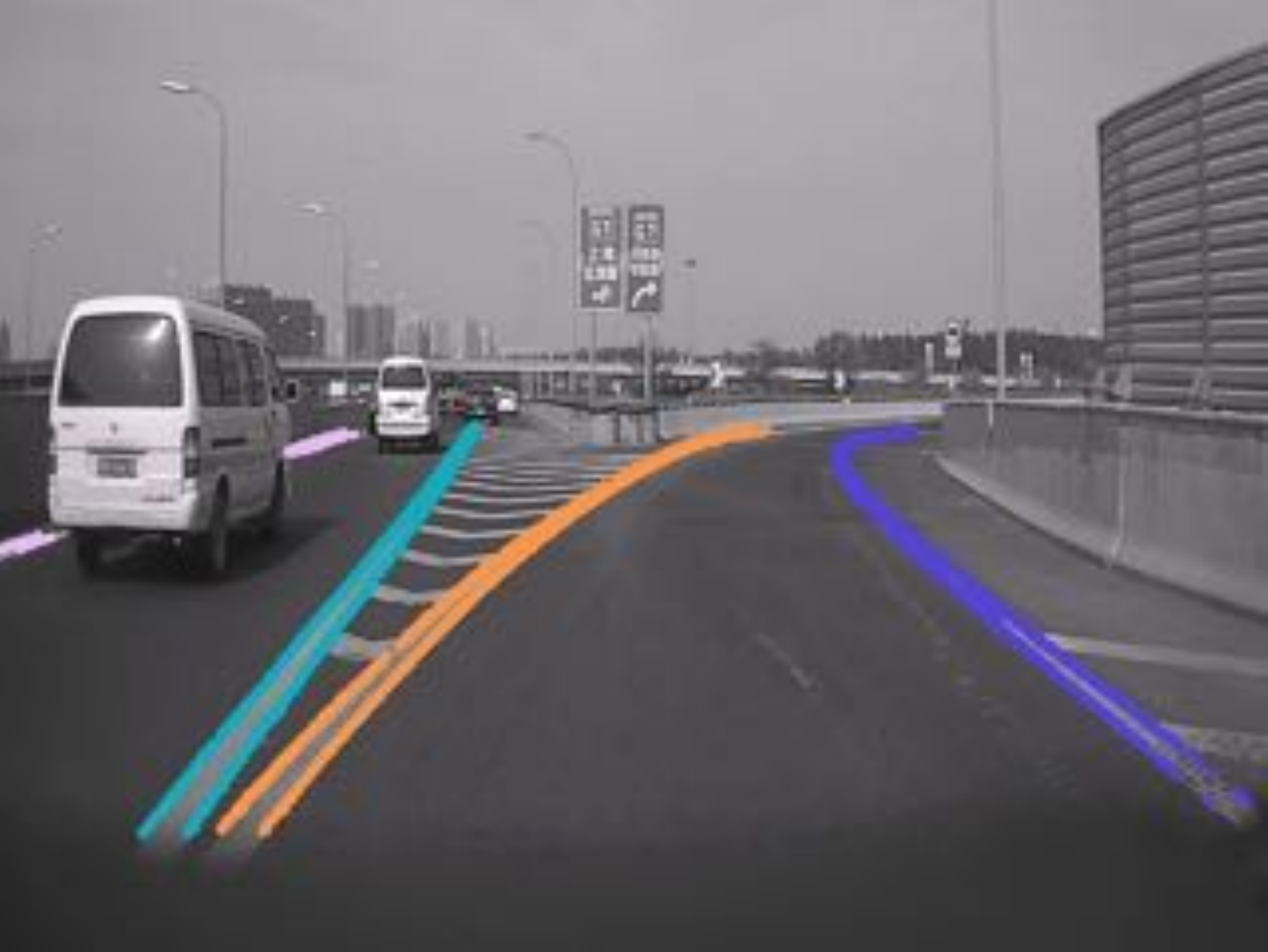}\vspace{5pt}
			\includegraphics[width=1\linewidth]{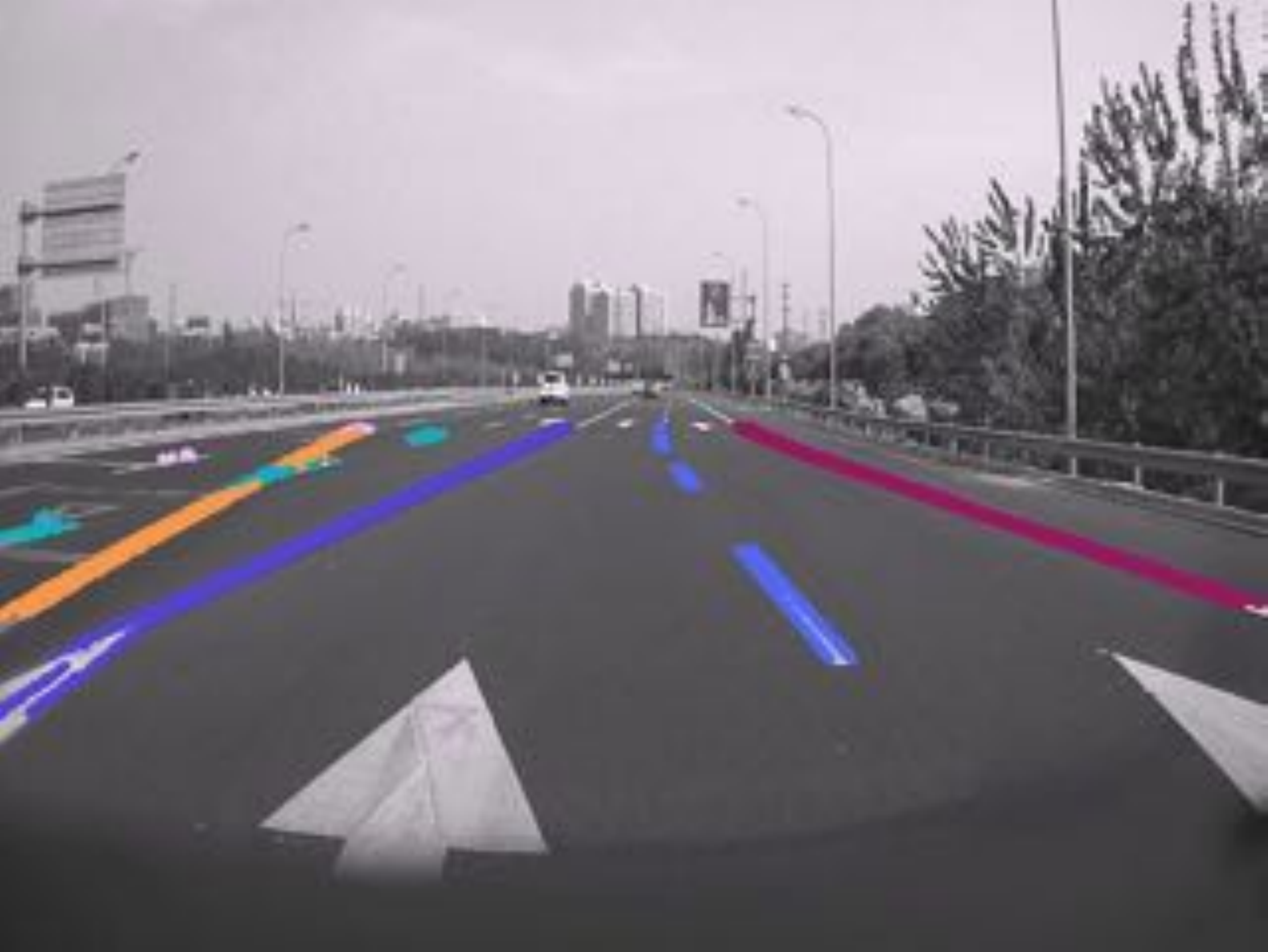}\vspace{5pt}
			\includegraphics[width=1\linewidth]{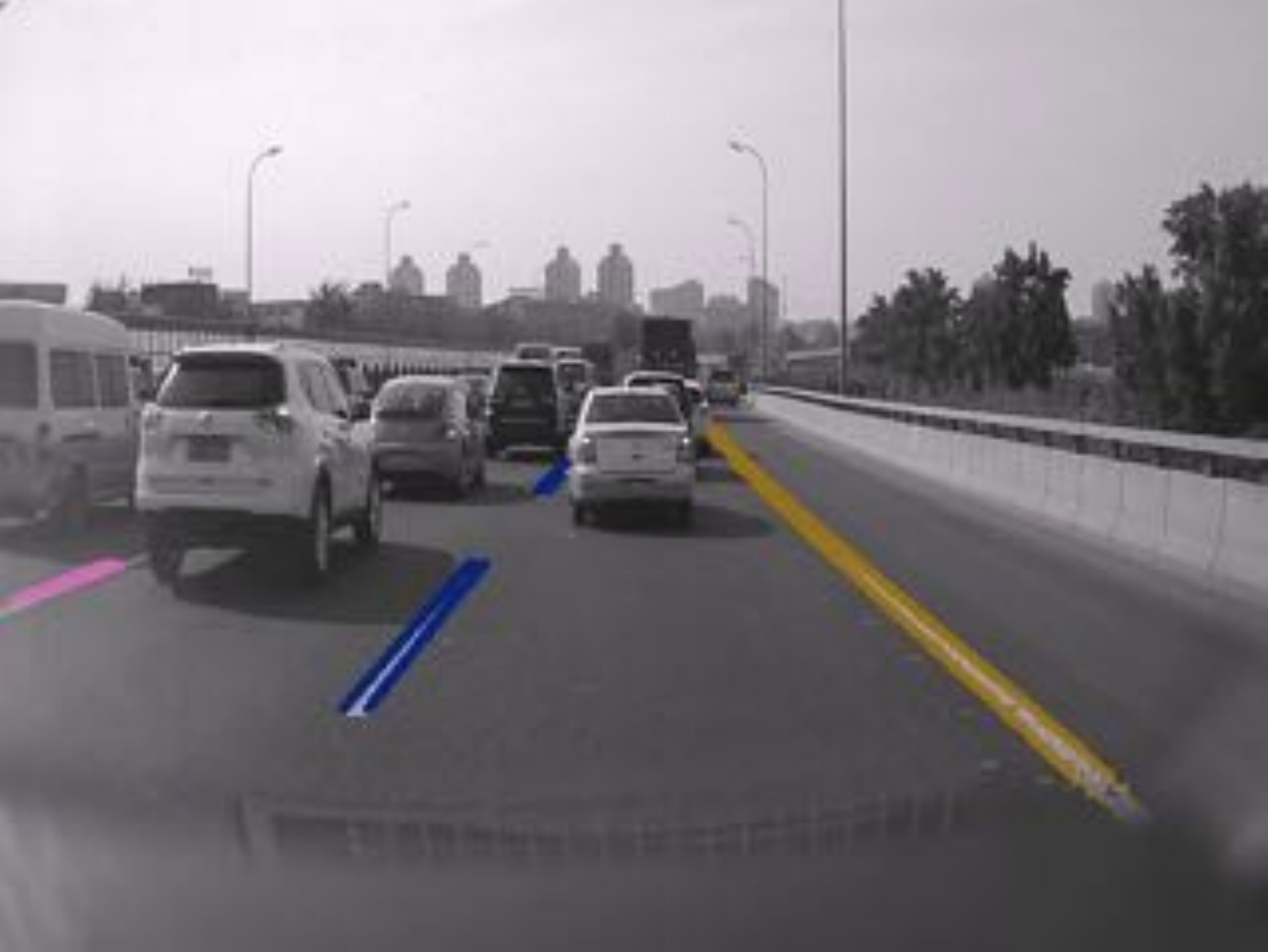}
	\end{minipage}}
	
	\caption{Illustrative results. The original images along with the corresponding lane edge proposal maps and final detection results are presented. In the final results, different lanes are marked in different colours. 
	}
	\label{fig:re}
\end{figure*}
\section{Experiments}
\label{sec:exp}

We test our LaneNet on the real world traffic data. In this section, we compare our method with other lane detection method, and evaluate the performance. Our data consist of more than 5,000 annotated front view images taken on both highways and urban roads. For each image in the dataset, we annotate all the edges of lane segments, and fit each lane line by a quadratic function.
We sample 600 samples from the dataset to be the test data and use the remaining data to train our LaneNet. Specifically, the test set is split into two sub-sets according to the difficulty.

We follow the widely used evaluation metric and adopt two criteria: 
the true positive rate (TPR) which is calculated as TPR = (the number of detect lanes) / (the number of target lanes), and false positive rate (FPR) which is calculated as FPR = (the number of false positives) / (the number of target lanes).
Each lane should be only detected once, so that both over estimation and under estimation of the total number of lanes are undesirable, e.g. detecting a dashed lane for multiple times results in a high FPR, and detecting splitting lanes to be one lane will lower the TPR.
We select the lane detection method proposed in \cite{Hur2013Multi} to compare the performance since it also relies on few assumptions, and the code is publicly available \footnote{We use the public version of the code which is not exactly identical with the original implementation as noted in the code.}. Table~\ref{tab:exp} shows the results.
The performance of our LaneNet is reported under the following training setting in terms of supervisions: lane edge proposal network is trained using the entire training dataset with full supervision. Weak supervision is introduced in the training of lane line localization network. Additional weakly labelled training samples (about 1,000 samples annotated by only the number of lane lines in the images) are used for fine tuning the parameters of lane line localization network.

\begin{table*}
	\begin{center}  
		\caption{Experiment results.}
		\label{tab:exp}
		\begin{tabular}{r|ccc|ccc}
			\toprule
			Difficulty & \multicolumn{3}{c|}{Easy (1170 lanes)} & \multicolumn{3}{c}{Hard (1035 lanes)} \\
			& Detected & TPR & FPR  & Detected & TPR & FPR \\
			\midrule
			
			Junhwa \textit{et al.} \cite{Hur2013Multi}  & 916 & 78.2\% & 9.5\% & 801 & 77.4\% & 15.9\% \\
			
			\textbf{LaneNet} & 1146	& 97.9\% & 2.7\% & 1001 & 96.7\% & 3.9\% \\
			\bottomrule
		\end{tabular}
	\end{center}  
\end{table*}

We present some illustrative results in Fig.~\ref{fig:re}. All the predictions are transformed back to the original images from the IPM images We transform all the predictions on IPM images back to original images for clear illustrations. 
The result of lane edge proposal network is robust to other objects on the roads, e.g. the lane marks and the vehicles, which we attribute to the contextual information extracting ability of deep neural network based lane edge proposal network, e.g., for every example, there is no false positive detection on the arrows on the roadways, which indicates that our deep neural network based detection method makes decision base on large contextual area, so that is robust to objects with similar local appearances.
The localizations of the lanes on the roads are accurate regardless the shapes and the number of the lanes, which demonstrate the effectiveness of our lane localization network to diverse real-world scenarios. The dashed lane lines are properly estimated, which we believe it's attributed to the ability of our lane localization network to evaluate the lane edge proposals from a global view instead of simply clustering the nearby points. 

We also conduct a additional experiment on validating the performance improvements by applying the weakly supervised loss function and additional weakly labelled samples. Results are presented in Table~\ref{tab:add}.
We can see that the weak supervision loss consistently improves the detection on both easy and hard sub-datasets.
Further fine tuning the network using weakly labelled data improves the performance on samples with hard difficulties significantly.
The fine tuning using weakly labelled data actually enables us to improve the network performance on extreme cases with low cost. We believe that more carefully collected hard samples will further improve the network to a even better stage.

\begin{table*}
	
	\begin{center}  
		\caption{Performance improvements by the weakly supervised loss function and additional weakly labelled samples.}
		\label{tab:add}
		\begin{tabular}{r|ccc|ccc}
			\toprule
			Difficulty & \multicolumn{3}{c|}{Easy (1170 lanes)} & \multicolumn{3}{c}{Hard (1035 lanes)} \\
			& Detected & TPR & FPR  & Detected & TPR & FPR \\
			\midrule
			
			Baseline   & 996 & 85.1\% & 3.4\% & 904 & 87.3\% & 7.7\% \\
			
			Weak supervision & 1129	& 96.4\% & 2.8\% & 932 & 90.0\% & 5.1\% \\
			Weak supervision + Tuning on weakly labelled data & 1146	& 97.9\% & 2.7\% & 1001 & 96.7\% & 3.9\% \\
			\bottomrule
		\end{tabular}
	\end{center}  
\end{table*}



Additionally, when using a NVIDIA Titan Xp GPU, our lane edge proposal network runs at the speed of 330 frame per second (FPS), and the lane line localization network is even 4 times faster, which enable our entire LaneNet to process images at the speed of 250 FPS. When running on an embedded GPU platform, e.g. NVIDIA Jetson TX1, the speed turns to 26 FPS without specific modifications, which is fast enough for real-time detection. The model size of the entire LaneNet is restricted to less than 1GB. Both the Compact model size and the high running speed further enable the deployment of our LaneNet on vehicles.

\section{Related work}
\label{sec:rw}

As a key component of autonomous driving system, lane detection have been widely studied for years. 
Despite the intensive demands of real world applications, lane detection remains challenging.
One of the key reasons is the lack of distinct features.
To solve this, many works seek to leverage the structural feature of lanes.
Vanishing point is considered to be an important feature for the detection of lanes and is used in \cite{lee2017vpgnet,ozgunalp2017multiple,su2017vanishing}.
As for local features, gradient on the lane edges is a popular tool for locating the lanes, e.g. in \cite{Hur2013Multi}, two filters which detect left and right edge, respectively, are adopted for finding the strong lane edges in the images.
However, gradient based detection methods are vulnerable to scenarios with complex irrelevant objects. 
From our observation, the performances of gradient based methods drop dramatically when large amount of other vehicles appear in the images, which are unavoidable in real world scenarios.

In recent years, with the fast development of deep learning and the hardware support, a lot of lane detection methods based on deep neural networks have emerged and show excellent performances since the training of deep neural network enables detection methods to learn features that are far more robust than the hand-crafted features. Alexandru Gurghian \textit{et al.} propose DeepLanes which use deep neural network to estimate lane position on the both side of the vehicle. Jun Li \textit{et al.} use a multitask convolutional neural network to simultaneously detect the presences and the geometric attributes of the lane marks in a patch of the image, however, the patch based detection makes it hard to infer the global structure of the lane lines on the road, thus is incapable of offering decisive information for navigation, so they propose to further combine the convolutional neural networks with recurrent neural networks to introduce context information and make structural predictions. Similar to our method, Xue Li \textit{et al.} propose using neural networks to detect the edge of lane lines, and use Hough transform to detect different lanes which the robustness might be weak when encountering with complicated scenarios like extremely curved lane lines. 

A more comprehensive review on recent advances in lane detection can be found in \cite{narote2018review}.
\section{Conclusion}
\label{sec:conclu}
In this paper, we proposed a lane detection method consists of two deep neural networks. The lane edge proposal network takes an IPM image of the front view of a vehicle as input and produces a lane edge proposal map. Given the lane edge map, the lane line localization network is then in charge of inferring the location of each lane in the image. Our lane detection method does not rely on any assumptions and can be applied to various situations.
Using deep neural network endows our method with great robustness and the two stages detection pipeline reduces the computational cost and allows our lane line localization network to be trained in a manner which combines supervised and weekly supervised learning, this remarkably reduces the cost of labelling training data. Extensive experiments demonstrate the speed, accuracy, and robustness of our LaneNet to diverse scenarios. 
The future work will focus on introducing lane tracking to our LaneNet for a more stable video based lane line detection method.
\newpage

{\small
\bibliographystyle{ieee}
\bibliography{egbib}
}

\end{document}